\documentclass{article}

\usepackage{microtype}
\usepackage{graphicx}
\usepackage{subfigure}
\usepackage{booktabs} %
\usepackage{float}
\usepackage{subfigure}

\usepackage[inline]{enumitem}

\usepackage{hyperref}

\usepackage[accepted]{icml2024}

\usepackage{amsmath}
\usepackage{amssymb}
\usepackage{mathtools}
\usepackage{amsthm}

\usepackage[capitalize,noabbrev]{cleveref}

\theoremstyle{plain}

\theoremstyle{definition}

\theoremstyle{remark}

\usepackage[textsize=tiny]{todonotes}

\icmltitlerunning{Unified Uncertainties: Combining Input, Data and Model Uncertainty into a Single Formulation}

\begin{document}
    
    \twocolumn[
    \icmltitle{Unified Uncertainties: Combining Input, Data and Model Uncertainty into a Single Formulation}

    \begin{icmlauthorlist}
        \icmlauthor{Matias Valdenegro-Toro}{rug}
        \icmlauthor{Ivo Pascal de Jong}{rug}
        \icmlauthor{Marco Zullich}{rug}
    \end{icmlauthorlist}
    
    \icmlaffiliation{rug}{Department of Artificial Intelligence, University of Groningen, Groningen, The Netherlands}

    \icmlcorrespondingauthor{Matias Valdenegro-Toro}{m.a.valdenegro.toro@rug.nl}
    \icmlkeywords{Machine Learning, ICML}
    
    \vskip 0.3in
    ]

    \printAffiliationsAndNotice{}  %

    \begin{abstract}

        Modelling uncertainty in Machine Learning models is essential for achieving safe and reliable predictions. Most research on uncertainty focuses on output uncertainty (predictions), but minimal attention is paid to uncertainty at inputs. We propose a method for propagating uncertainty in the inputs through a Neural Network that is simultaneously able to estimate input, data, and model uncertainty. Our results show that this propagation of input uncertainty results in a more stable decision boundary even under large amounts of input noise than comparatively simple Monte Carlo sampling. Additionally, we discuss and demonstrate that input uncertainty, when propagated through the model, results in model uncertainty at the outputs. The explicit incorporation of input uncertainty may be beneficial in situations where the amount of input uncertainty is known, though good datasets for this are still needed.

    \end{abstract}
    
    \section{Introduction}
    
    While neural networks are state of the art for many problems ranging from language generation to image interpretation, their predictions are often overconfident and they have issues estimating their own uncertainty \cite{valdenegro2022deeper, ovadia2019can}.
    
    One often overlooked source of uncertainty is the model's input \cite{tzelepis2017linear}, as features and pixel values can be noisy, depending on the data source. Uncertainty in the input is an underexplored research area \cite{rodrigues2021information, hullermeier2014learning, depeweg2018decomposition}, with most works about uncertainty estimation being about output uncertainty \cite{hullermeier2021aleatoric}. Accounting for the uncertainty in the input can improve the final prediction, as well as its uncertainty. Appendix \ref{sec:related-work} offers further discussion on related work.
    
    In this work we introduce the concept of input uncertainty, as well as a new formulation for simultaneously estimating data uncertainty (aleatoric---AU), model uncertainty (epistemic---EU) and input uncertainty---IU. Additionally, we argue that propagation of IU into the model results in epistemic uncertainty. 
        
    The contributions of this work are: 
    \begin{enumerate*}[label=(\alph*)]
        \item a novel formulation combining AU, EU, and IU into a single model and uncertainty estimation method,
        \item a theoretically grounded interpretation that IU is processed as EU, and
        \item an experimental validation of our formulation.
    \end{enumerate*}

    \begin{figure}
        \centering
        \subfigure[$\sigma = 0.10$]{\includegraphics[width=0.35\linewidth]{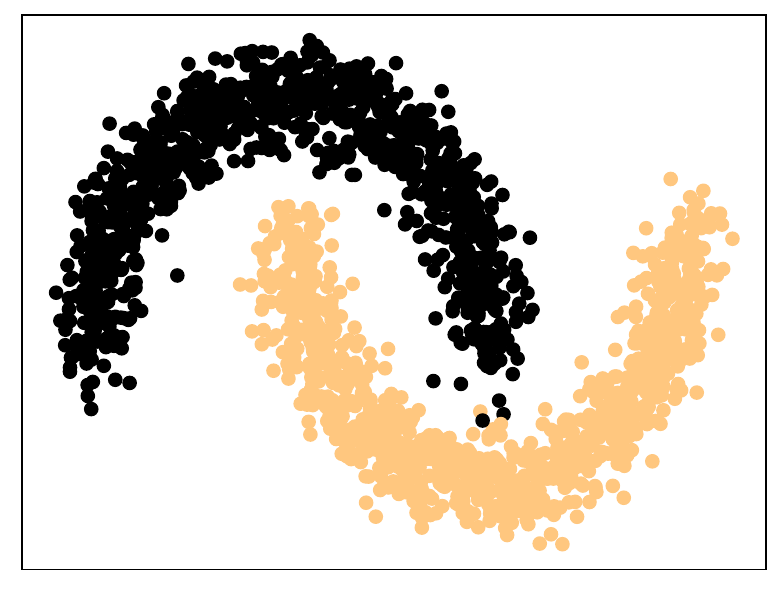}}
        \subfigure[$\sigma = 0.25$]{\includegraphics[width=0.35\linewidth]{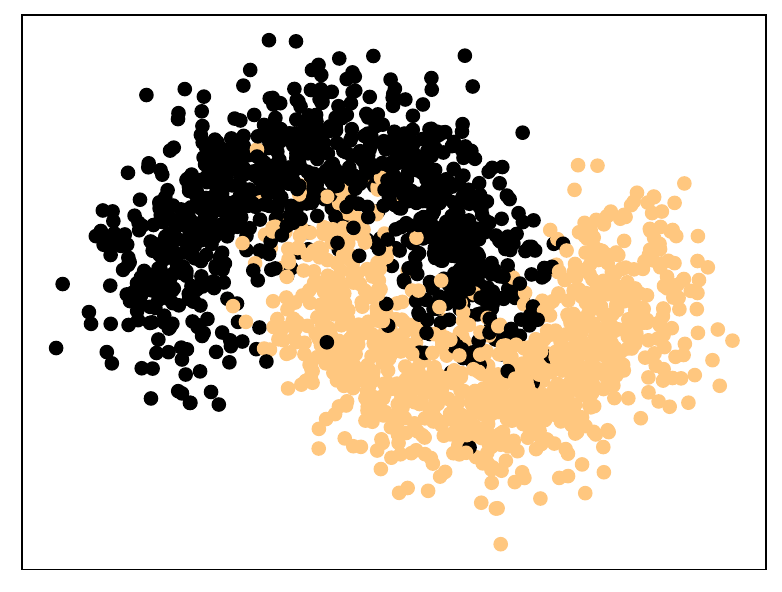}}
        \vspace{-0.7em}
        \caption{Concept of Input Uncertainty in the Two Moons dataset, increasing $\sigma$ makes classification more difficult and decision boundary unclear. Considering IU could improve model performance.}
    \end{figure}
    
    \section{Input Uncertainty Formulation}
    
    \textbf{Notation}. We denote standard supervised learning datasets as $(\mathbf{x}_j, y_j)_{1}^{N}$. Datasets can have IU, which is denoted as $\mathbf{x} = \{ \mu^i, \sigma^i \}$ with $^i$ denoting input variables. Datasets can additionally have output uncertainty, which is denoted differently for regression and classification. For regression $y = \{ \mu^o, \sigma^o \}$, with $^o$ denoting output variables. For classification $y = P(\hat{y} {=} c \,|\, \mathbf{x}) \quad \forall c \in \{1,\dots,C\}$, where probabilities can be obtained via sampling through the softmax function $P(\hat{y} {=} c \,|\, \mathbf{x}) \sim \text{softmax}(z_c)$ and \mbox{$z_c \sim \mathcal{N}(\hat{\mu}^o, \hat{\sigma}^o)$}. A model is a function $f_\theta$ with trainable parameters $\theta$. Ground truth labels are denoted as $y$, while predicted values are $\hat{y}$.

    We define the input/output uncertainty estimation problem as training a model $f_\theta$ where both IUs $\sigma^i$ and output uncertainties $\sigma^o$ are considered, and related by:
    \begin{equation}
        \hat{y}^o = f_\theta(\mathbf{x}^i)
    \end{equation}
    A common desideratum is that output uncertainties are calibrated.
    To denote inputs and outputs with uncertainty, we use the notation $\mu \pm \sigma$, where $\mu$ is the prediction or input value and $\sigma$ is its associated uncertainty, represented by the standard deviation.
    
    \textbf{IU Propagation}. Given an input with uncertainty \mbox{$x = \{ \mu^i, \sigma^i \}$}, it is possible to propagate the uncertainties through a model $f_\theta$ by using a first order Taylor approximation \cite{kelly19943d}:
    \begin{equation}
        f_\theta(\mu^i \pm {\sigma^2}^i) \approx f_\theta(\mu) \pm \mathbf{J} \sigma^2 \mathbf{J}^T
        \label{eq:iu_propagation}
    \end{equation}
    Where $\mu^o = f_\theta(\mu^i)$ and ${\sigma^o}^2 = \mathbf{J} {\sigma^i}^2 \mathbf{J}^T$ and the Jacobian matrix is $\mathbf{J}_{ij} = \frac{\partial f_i}{\partial \mu_j}$ evaluated at $\mu^i$.
    
    \textbf{IU MC Sampling}. An alternative way to propagate IU to the output of a model $f_\theta$ is via Monte Carlo sampling:
    \begin{equation}
        f_\theta(\mu^i \pm {\sigma^2}^i) \approx \mathbb{E}[f_\theta(\tilde{\mathbf{x}})] \pm \text{Var}[f_\theta(\tilde{\mathbf{x}})] \quad \tilde{\mathbf{x}} \sim \mathcal{N}(\mu^i, {\sigma^2}^i)
        \label{eq:iu_mc}
    \end{equation}
    The input is modelled as Gaussian random variable $\tilde{\mathbf{x}}$, from which samples are drawn that are passed through the model $f_\theta$. The output is an approximate Gaussian $\mathcal{N}(f_\theta(\tilde{x}), \text{Var}[f_\theta(\tilde{x})])$. The monte carlo approximation is computed using $N$ samples:
    \begin{align*}
        \mathbb{E}[f_\theta(\tilde{\mathbf{x}})] &\approx N^{-1} \sum^N f_\theta(\tilde{\mathbf{x}}_j)\\
        \text{Var}[f_\theta(\tilde{\mathbf{x}})] &\approx (N-1)^{-1} \sum^N (f_\theta(\tilde{\mathbf{x}}_j) - \mathbb{E}[f_\theta(\tilde{\mathbf{x}})])^2
    \end{align*}
    With $\tilde{\mathbf{x}}_j$ being the j-th sample from $\mathcal{N}(\mu^i, {\sigma^2}^i)$. Both approaches are approximate, as there is no analytic formulation for an exact propagation of uncertainty.
    
    \subsection{Uncertainty Estimation Formulation}
    
    In this section we combine IU with standard formulations for AU and EU. For IU estimation, we assume the existence of a function $\text{\texttt{propagate}}(\mu, \sigma^2)$ that propagates IU through the model, as described in the previous section, and $\texttt{epistemic}(\mu)$ corresponds to standard uncertainty estimation methods that produce EU via sampling (e.g. MC-Dropout \cite{gal2016dropout}, MC-DropConnect \cite{mobiny2021dropconnect}, Bayesian Neural Networks like Flipout \cite{wen2018flipout}, etc) or ensembles (e.g. Deep Ensembles \cite{lakshminarayanan2017simple}). We combine these by first applying the EU estimation method, and then the IU propagation method:
    \begin{align*}
        S_\text{epi}(\mu, \sigma^2) &= \texttt{propagate}(\texttt{epistemic}(\mu), \sigma^2)\\
        \mu^s &= \mathbb{E}[S_\text{epi}(\mu, \sigma^2)]\\
        {\sigma^2}^s &= \text{Var}[S_\text{epi}(\mu, \sigma^2)]
    \end{align*}
    Where $S_\text{epi}(\mu, \sigma^2)$ represents the result of the IU propagation method, corresponding to $N$ samples (on which expectation and variance are computed), depending on the IU propagation method, while the EU method uses $M$ forward passes. Then three values can be estimated:
    \begin{align*}
        \mu^o &= \mathbb{E}[\mu^s] &\quad \quad  (\text{Prediction})\\
        \sigma_{\text{inp}}^o &= \text{Var}[\mu^s] &\quad \quad (\text{IU})\\
        \sigma_{\text{epi}}^o &= \mathbb{E}[{\sigma^2}^s] &\quad \quad (\text{EU})
    \end{align*}
    In general terms, $\mu^o$ is the model prediction after averaging over the EU estimation method, $\sigma_{\text{inp}}$ is the output uncertainty attributed to the input, being the disagreement or variation between the samples, and $\sigma_{\text{epi}}$ is the output EU, given by the average of the per-input-sample EUs produced by $\texttt{epistemic}(\mu)$.
    
    Depending on the task, the prediction also holds AU.
    
    \textbf{Classification}. For classification with $C$ classes, output uncertainties ($\mu^o$, $\sigma_{\text{inp}}$,  $\sigma_{\text{epi}}$) are modelled as logits (values before applying softmax), so they are C-dimensional vectors that can be transformed into probabilities by:
    \begin{align*}
        p_{\text{ale}}(y \,| \, \mathbf{x}) &= \text{softmax}(\mu^o) &\quad (\text{AU})\\
        p_{\text{inp}}(y \,| \, \mathbf{x}) &= \text{sampling\_softmax}(\mu^o, \sigma_{\text{inp}}) &\quad (\text{IU})\\
        p_{\text{epi}}(y \,| \, \mathbf{x}) &= \text{sampling\_softmax}(\mu^o, \sigma_{\text{epi}}) &\quad (\text{EU})
    \end{align*}
    Where the sampling softmax function \cite{valdenegro2022deeper} is defined by:
    \begin{align*}
        \text{sampling\_softmax}(\mu(\mathbf{x}), \sigma^2(\mathbf{x})) &= N^{-1} \sum_j \text{softmax}(\hat{\mathbf{z}}_j)\\[0.1em]
        \text{with } \hat{\mathbf{z}}_j &\sim \mathcal{N}(\mu(\mathbf{x}), \sigma^2(\mathbf{x}))
        \addtocounter{equation}{1}\tag{\theequation}
    \end{align*}
    Where $N$ is the number of samples used for Monte Carlo sampling.
    
    \textbf{Regression}. Regression requires small changes to the formulation. To estimate AU in regression, a model with two output heads is required, and the Gaussian negative log-likelihood loss \cite{seitzer2022on} is used to train both mean $\mu$ and variance $\sigma^2$ heads.
    \begin{equation}
        L_{NLL}(y_j, \mathbf{x}_j) = \log \sigma^2(\mathbf{x}_j) + \frac{(\mu(\mathbf{x}_j) - y_j)^2}{ \sigma^2(\mathbf{x}_j)} .
        \label{eq:gaussian_nll}
    \end{equation}
    The model's forward passes now produce mean and variance, implying that $\texttt{epistemic}(\mu)$ now returns two values. Some adjustments are needed for the regression formulation, to propagate aleatoric variances from the model forward passes through the epistemic and IU propagation methods, without changes, only producing samples.
    \begin{equation}   
        S_\text{epi}(\mu, \sigma^2), \sigma^2_\text{ale} = \texttt{propagate}(\texttt{epistemic}(\mu), \sigma^2)
    \end{equation}
    Then the associated predictions and uncertainties for regression are:
    \begin{align*}
        \mu^o, \sigma_{\text{ale}}^o &= \mathbb{E}[\mu^s], \mathbb{E}[{\sigma^2}_\text{ale}] &\quad \quad (\text{Prediction \& AU})\\
        \sigma_{\text{inp}}^o &= \text{Var}[\mu^s] &\quad \quad (\text{IU})\\
        \sigma_{\text{epi}}^o &= \mathbb{E}\big[{\sigma^2}^s\big] &\quad \quad (\text{EU})
    \end{align*}
    The output aleatoric variance $\sigma_{\text{ale}}^o$ is the expectation over aleatoric uncertainty samples propagated through both $ \texttt{propagate}()$ and $\texttt{epistemic}()$.
    
    \subsection{Discussion}
    \textbf{Uncertainty Interpretation}. Conceptually the IU $\sigma^i$ is part of AU, as its source is the training data and the sensor or device producing the original data. IU does not reduce if more information or data is added, as it is inherent to the data. %
    
    When propagating IU through a model, the output uncertainties $\sigma_{\text{inp}}^o$ and $\sigma_{\text{epi}}^o$ are EUs, as they correspond to uncertainty due to the model's equations, and it is combined with an EU estimation method. This is a novel view as we argue that input AU is transformed into output EU via our proposed uncertainty formulation.
    The predictive uncertainty can be conceptually combined as:
    \vspace*{-0.2cm}
    \begin{equation}
        \footnotesize
        \text{Predictive Uncertainty} = \text{Aleatoric} + \text{Epistemic} + \text{Input}
    \end{equation}
    \textbf{Computational Costs}. Methods for IU estimation have increased computational requirements during inference. For IU propagation using a Taylor approximation, the cost is $\mathcal{O}(M)$, as only EU estimation contribute to increased costs, but sampling IU estimation has cost $\mathcal{O}(NM)$, due to $N$ samples for IU propagation, and each of them needs $M$ samples/ensembles for EU estimation. This makes sampling infeasible on large models and datasets, while using the Taylor approximation incurs minimal additional cost. 
    
    \section{Experiments}      
    
    \textbf{Experimental Setup}. For experimentation we use the two moons dataset because it is simple, we can add noise to input data points to simulate uncertainty, and it allows a clear visualization of uncertain regions and the decision boundary. We train a four layer multi-layer perceptron using ReLU activations, the two last layers use some form of uncertainty estimation method as described below. Models are trained with the Adam optimizer using the cross-entropy loss.
    
    Gaussian distributed noise with $\mu = 0$ and $\sigma = 0.1$ is added to the training set, and for IU propagation evaluation, we use increasing levels of IU: $\sigma \in \{0.25, 0.5, 1.0\}$.
    
    \textbf{EU Methods}. We evaluate a standard neural network (without EU methods), MC-Dropout \cite{gal2016dropout} with drop probability $p = 0.2$, MC-DropConnect \cite{mobiny2021dropconnect} with drop probability $p = 0.05$, Ensembles \cite{lakshminarayanan2017simple} with $M = 5$ ensemble members, and Flipout \cite{wen2018flipout}. Stochastic methods use $M = 20$ forward passes.
    
    \textbf{Expectations}. Validating IU propagation is difficult as there is no ground truth label for uncertainties, and novelty in formulations require a different evaluation approach. To evaluate the quality of IU propagation, we train a model with a certain level of IU on training data, and test model predictions and associated uncertainties with increasing levels of IU, expecting that the model will output increasing uncertainties, in particular $\sigma_{\text{inp}}^o$ should increase together with the IU, while AUs should remain approximately constant. EU should also increase but not at the same level as $\sigma_{\text{inp}}^o$.
    
    \subsection{Classification on Two Moons Dataset with IU}
    
\begin{figure*}[t]
        \centering
        \begin{minipage}{\textwidth}
            \centering
            \includegraphics[width=0.94\linewidth]{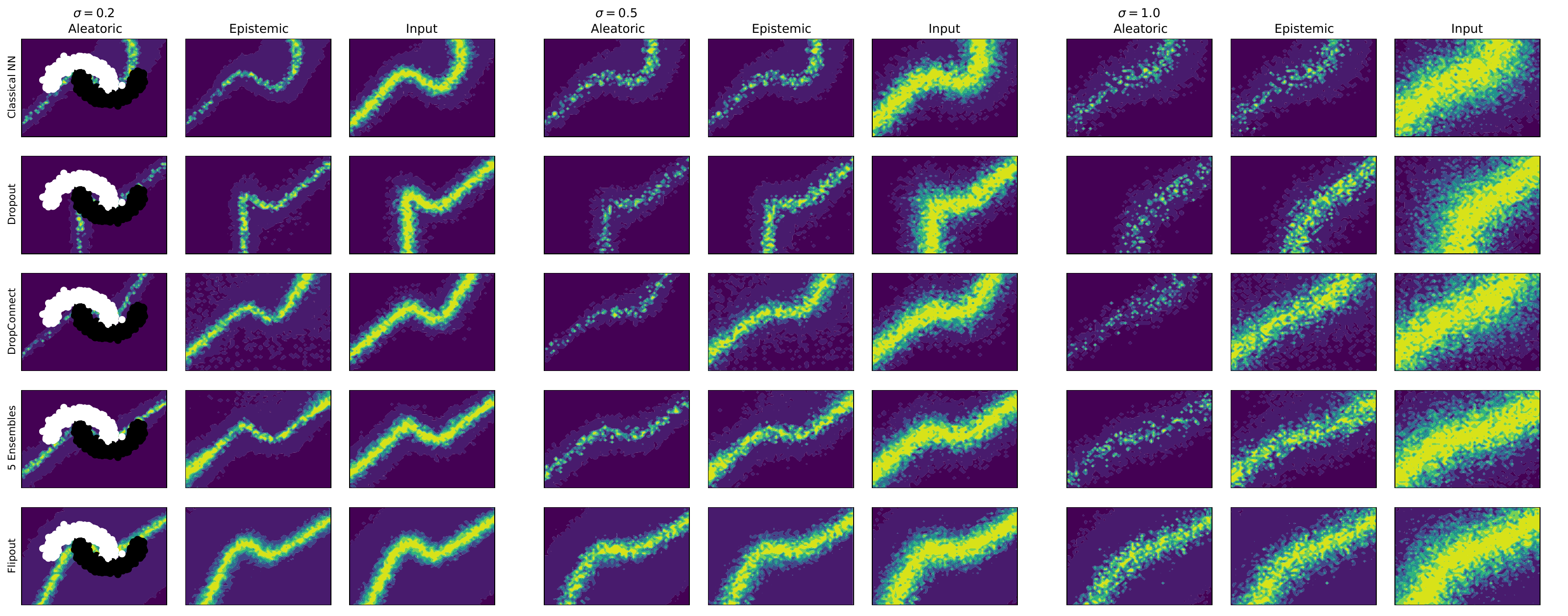}
            \caption{Results on the Two Moons dataset with Monte Carlo Sampling IU (Using Eq \ref{eq:iu_mc}). As the IU increases, the learned decision boundary loses shape and becomes noisy. Note that each value of $\sigma$ produces three plot columns for each output uncertainty.}
            \label{fig:sampling_results}
        \end{minipage}
        \vspace{-1em}
        \begin{minipage}{\textwidth}
            \centering
            \includegraphics[width=0.94\linewidth]{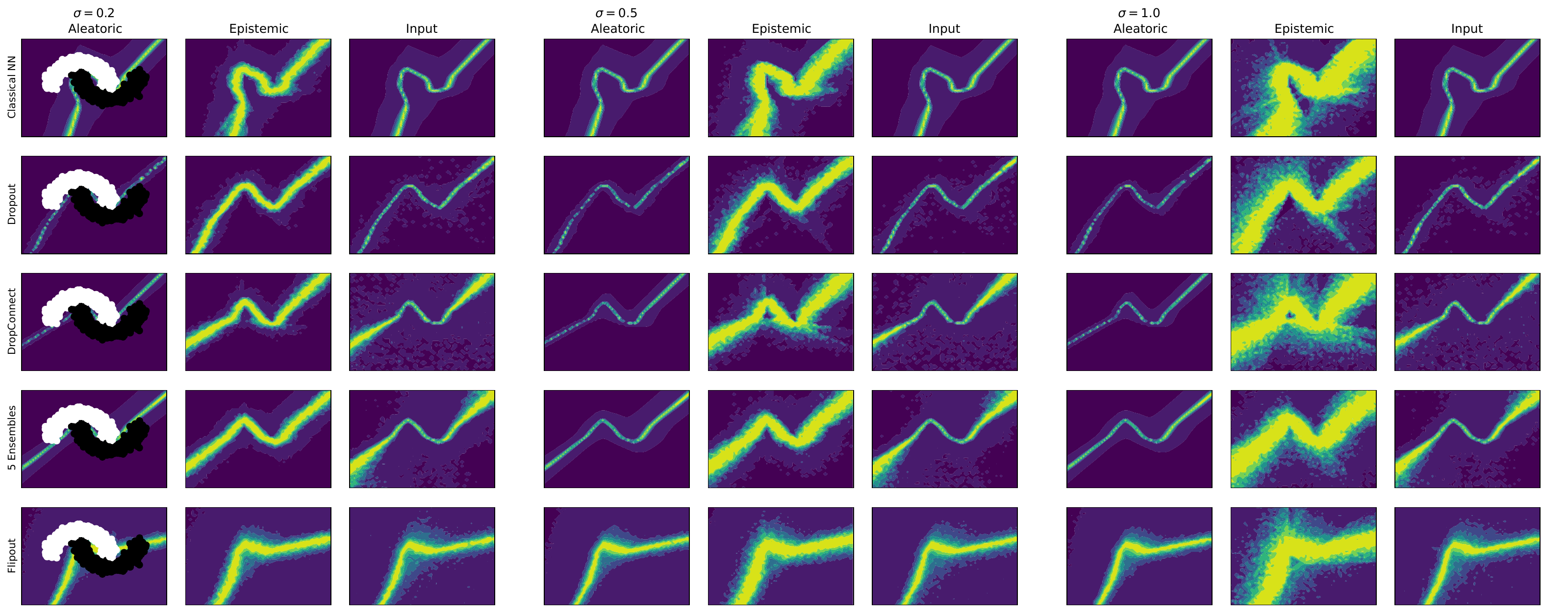}
            \caption{Results on the Two Moons dataset with Propagation IU (Using Eq \ref{eq:iu_propagation}). As the IU increases, the predicted EU increases. The other predicted uncertainties remain roughly the same. Note that each value of $\sigma$ produces three plot columns for each output uncertainty.}
            \label{fig:propagation_results}
        \end{minipage}        
    \end{figure*}

    Figures \ref{fig:sampling_results} and \ref{fig:propagation_results} demonstrate predicted AU, EU, and IU under different amounts of IU using respectively the Monte Carlo Sampling approach and the Uncertainty Propagation approach across different methods for quantifying EU. 
    The most salient difference between these two approaches, which seems consistent regardless of the method for quantifying EU, is that the Propagation approach maintains a clear decision boundary under increased input noise, while the Monte Carlo Sampling approach shows a decision boundary that becomes noisy and loses shape under high amounts of input noise. This suggests propagation of input uncertainty may affect the decision boundary. 
                
    The second difference is that under the Uncertainty Propagation approach, the predicted IU changes very little, but the predicted EU increases. This effect also exists in the Monte Carlo approach, but is less pronounced. The propagation of IU leads to an increase in EU as it offers the necessary information so that a better decision boundary may be learned, but this does add complexity to the learning problem.
    
    Overall, the Monte Carlo Sampling shows that the increase in IU adds noise to all predicted uncertainties. However, for the Uncertainty Propagation, only the EU changes. This shows that the Uncertainty Propagation approach allows for a more consistent disentanglement of the different kinds of uncertainty, and allows the preservation of decision boundaries even under large amounts of input noise. 
    
    The main drawback of the Uncertainty Propagation approach is that it fails to predict an increase in IU even though the ground truth IU does increase. The Sampling IU does predict an increase in IU.
    
    The different methods for predicting EU exhibit similar behaviour, showing that IU has a minor role in direct epistemic uncertainty, clearly being reflected via the method used to propagate IU through the model. Appendix \ref{sec:regression_results} shows that the findings for classification also hold for regression.
    
    \section{Conclusions and Future Work}
    
    In this work we proposed a new formulation for unified uncertainties considering aleatoric, epistemic, and input uncertainty. This allows for input uncertainty to be explicitly modelled in a neural network.
    
    The results show that the proposed Uncertainty Propagation method results in better uncertainty estimates and more consistent decision boundaries than a comparatively simple baseline method of Monte Carlo Sampling from the input uncertainty. Input uncertainty is often trivialised as noise that can just be learned with sufficient samples. Our work shows that knowledge about the input uncertainty may be propagated effectively in a Neural Network resulting in more desirable predictive behaviour of the model. Additionally, we show a new insight that this aleatoric input uncertainty becomes epistemic uncertainty in the output if propagated through the model. 
    
    The experiments with toy data and the underlying theory show that this is a promising new avenue for further research on uncertainty quantification. To really capitalise on input uncertainty these methods should be applied to cases where input uncertainty may be estimated from prior knowledge. Depth-sensing cameras such as Kinect for example are known to have a higher depth-variance on object further away, and many sensors are manufactured with known amounts of (im)precision in their observation. Collecting this kind of information in datasets is the next step in exploring the viability of propagating input uncertainty through a Neural Network.        
    
    \clearpage
    \bibliography{biblio}
    \bibliographystyle{icml2024}

    \newpage
    \appendix

    \section{Broader Impact Statement}
    Machine learning models and systems using them are generally not robust to different types of corruptions in the input data, and we believe that explicitly modelling uncertainty and noise in the input data can lead to more robust models, as more information is provided to make a prediction.

    But as with any uncertainty estimation method, there are no guarantees about correctness of predictions and their associated uncertainties, so these methods should be used with care and extensive validation.

    Some limitations of our work are: we experiment only on toy datasets to understand how methods are actually working, and real-world performance is left for future work. We only test with input Gaussian noise, while real-world corruptions might follow distributions that considerably deviate from Gaussian noise.
    
    \section{Related Work}\label{sec:related-work}
    Most existing research on Uncertainty Quantification focuses on AU \cite{seitzer2022pitfalls}, EU \cite{gal2016dropout, lakshminarayanan2017simple, wen2018flipout}, or both \cite{valdenegro2022deeper, wimmer2023quantifying}, but very little work focuses on IU.

    In the typical BNN-based setup AU is generally estimated by a predicted variance for regression \cite{seitzer2022pitfalls}, or by relying on the softmax function for classification \cite{guo2017calibration}. The desideratum in AU estimation is to have the predicted probability of a prediction match the probability of it being correct (accuracy). This is challenging under heteroscedastic aleatoric uncertainty, where the probability of being correct may change between different samples.     

    EU is modelled by uncertainty in the parameters. In true Bayesian Neural Networks each parameter in the Neural Network is represented with an arbitrary probability distribution. Learning this is computationally infeasible, so a variety of approximation methods have been proposed with varying estimation quality and practical consequences. MC-Dropout \cite{gal2016dropout} and MC-Dropconnect \cite{mobiny2021dropconnect} require minimal changes to the model architecture, and do not make any changes to the training process, but are $M$ times as computationally expensive during inference. Deep Ensembles \cite{lakshminarayanan2017simple} train $N$ models on the same data with different random initialization. This puts no requirements on the model architecture, and does not require as many forward passes to get good results, but makes training $N$ times more expensive and requires more memory to store the models. Flipout \cite{wen2018flipout} requires changes to the architecture, but does actually learn a distribution for the parameters.  

    Research on IU is limited and has not been connected with modern UQ literature. \citet{matsuoka1992noise} and \citet{sietsma1991creating} explored injecting uncertainty at the inputs as a form of regularization, but did not consider propagating IU to achieve better estimated uncertainties. Their findings input uncertainty regularizes can also be seen in Figure \ref{fig:sampling_results}.
    \citet{wright2000neural} and \citet{wright1998neural} does propagate IU through a Bayesian Neural Network for regression with the Laplace approximation. They show that the predicted uncertainty increases corresponding with the input uncertainty, as is also shown in our results. 

    \citet{bocsi2013hessian} proposes using the Hessian to correct biases when integrating input noise, corresponding to IU.

\section{Regression with IU on Toy Dataset} \label{sec:regression_results}

We additionally produce results on a toy regression problem, to test the equivalency of the regression formulation. We use a synthetic function corrupted with noise, defined by:
\begin{equation}
    f(x) = x \sin(x) + \epsilon_1 + \epsilon_2 x
\end{equation}
With $\epsilon_1 \sim \mathcal{N}(0, 0.3^2)$ and $\epsilon_2 \sim \mathcal{N}(0, 0.3^2)$. The use of both noise terms is to introduce homoscedastic ($e_1$) and heteroscedastic ($\epsilon_2$) aleatoric uncertainty to the data. No noise is added to the input variable $x$.

We generate $N = 1000$ equally spaced samples in $x \in [0, 10.0]$ for training the model, and additionally generate $N = 200$ samples in $x \in [10, 12]$ as out of distribution samples, to test behavior of input and epistemic uncertainty in this setting.

Input uncertainty is produced by corrupting the input with Gaussian noise with $\mu 0$ and increasing levels of noise following $\sigma \{ 0.25, 0.5, 1.0 \}$.

\begin{figure*}[t]
    \centering
    \includegraphics[width=\textwidth]{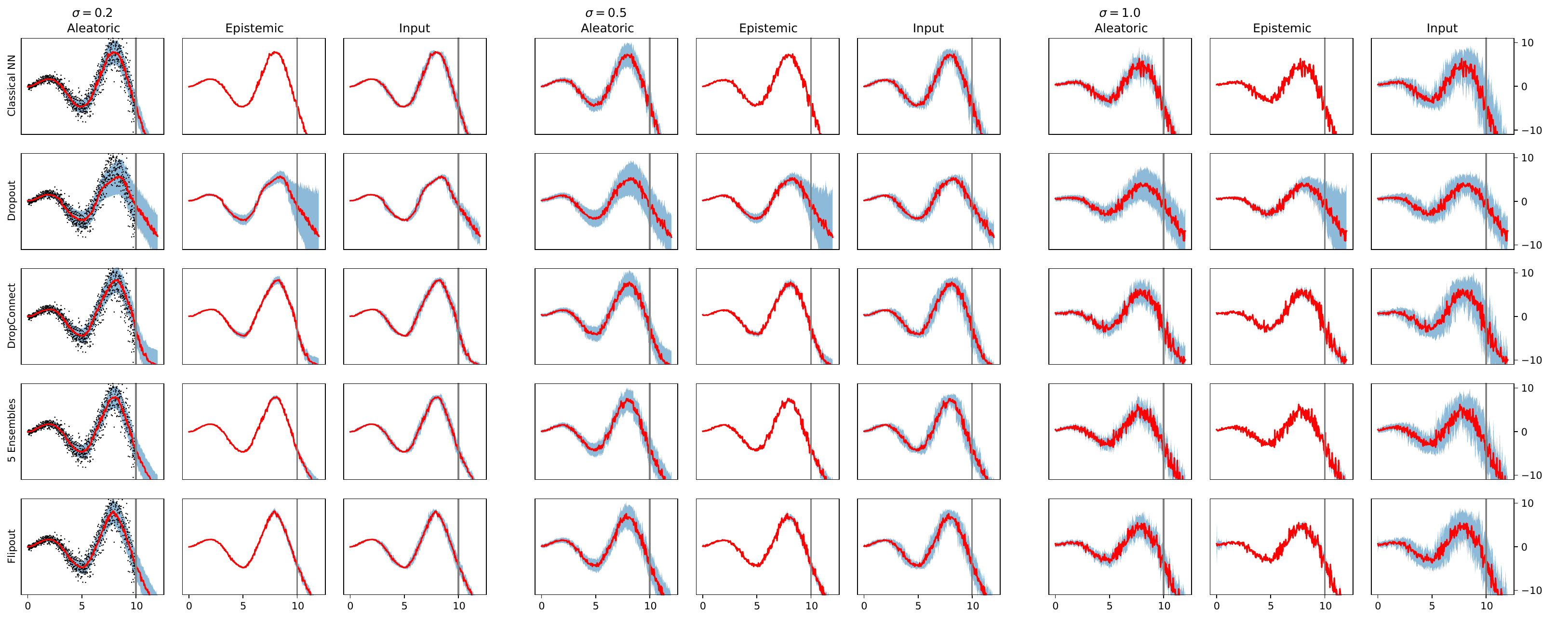}
    \caption{Results on the Toy Regression example with Monte Carlo Sampling IU. As the IU increases, predicted IU increases accordingly, without affecting EU. Note that each value of $\sigma$ produces three plot columns for each output uncertainty type (Aleatoric, Epistemic, Input).}
    \label{fig:reg_sampling}
\end{figure*}

\begin{figure*}[t]
    \centering
    \includegraphics[width=\textwidth]{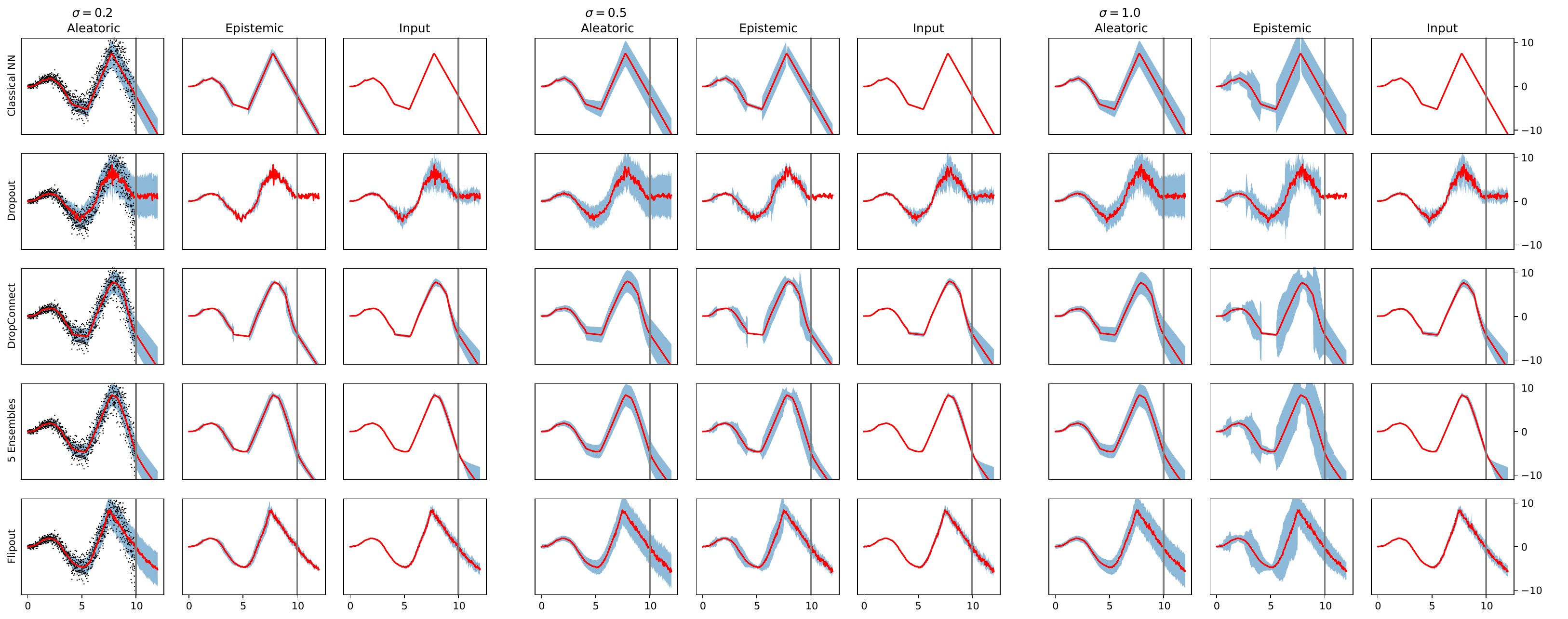}
    \caption{Results on the Toy Regression example with Propagation IU. As the IU increases, mostly predicted EU increases, while predicted IU increases only slightly. Note that each value of $\sigma$ produces three plot columns for each output uncertainty type (Aleatoric, Epistemic, Input).}
    \label{fig:reg_propagation}
\end{figure*}

Results for both sampling and propagation are presented in Figures \ref{fig:reg_sampling} and \ref{fig:reg_propagation} correspondingly. These results show similar behavior as with classification on two moons, with Monte Carlo sampling input uncertainty behaving as expected, while input uncertainty propagation turns input uncertainty into epistemic uncertainty, which is unexpected but in line with classification experiments. 

\end{document}